\documentclass[sigconf]{acmart}
\usepackage{multirow}

\AtBeginDocument{%
  }

\copyrightyear{2023}
\acmYear{2023}
\setcopyright{acmlicensed}\acmConference[MM '23]{Proceedings of the 31st ACM International Conference on Multimedia}{October 29-November 3, 2023}{Ottawa, ON, Canada}
\acmBooktitle{Proceedings of the 31st ACM International Conference on Multimedia (MM '23), October 29-November 3, 2023, Ottawa, ON, Canada}
\acmPrice{15.00}
\acmDOI{10.1145/3581783.3611750}
\acmISBN{979-8-4007-0108-5/23/10}

\begin{document}

\title{MVFlow: Deep Optical Flow Estimation of Compressed Videos with Motion Vector Prior}

\author{Shili Zhou}
\authornote{Both authors contributed equally to this research.}
\affiliation{%
  \institution{School of Computer Science, Shanghai Key Laboratory of Intelligent Information Processing, Shanghai Collaborative Innovation Center of Intelligent Visual Computing, Fudan University}
  \city{Shanghai}
  \country{China}
}
\email{slzhou19@fudan.edu.cn}

\author{Xuhao Jiang}
\authornotemark[1]
\affiliation{%
  \institution{School of Computer Science, Shanghai Key Laboratory of Intelligent Information Processing, Shanghai Collaborative Innovation Center of Intelligent Visual Computing, Fudan University}
  \city{Shanghai}
  \country{China}
}
\email{20110240011@fudan.edu.cn}

\author{Weimin Tan}
\authornote{Corresponding authors: Bo Yan, Weimin Tan. This work is supported by NSFC (Grant No.: U2001209) and Natural Science Foundation of Shanghai (21ZR1406600).}
\affiliation{%
  \institution{School of Computer Science, Shanghai Key Laboratory of Intelligent Information Processing, Shanghai Collaborative Innovation Center of Intelligent Visual Computing, Fudan University}
  \city{Shanghai}
  \country{China}
}
\email{wmtan@fudan.edu.cn}

\author{Ruian He}
\affiliation{%
  \institution{School of Computer Science, Shanghai Key Laboratory of Intelligent Information Processing, Shanghai Collaborative Innovation Center of Intelligent Visual Computing, Fudan University}
  \city{Shanghai}
  \country{China}
}
\email{rahe16@fudan.edu.cn}

\author{Bo Yan}
\authornotemark[2]
\affiliation{%
  \institution{School of Computer Science, Shanghai Key Laboratory of Intelligent Information Processing, Shanghai Collaborative Innovation Center of Intelligent Visual Computing, Fudan University}
  \city{Shanghai}
  \country{China}
}
\email{byan@fudan.edu.cn}

\begin{abstract}
  In recent years, many deep learning-based methods have been proposed to tackle the problem of optical flow estimation and achieved promising results. However, they hardly consider that most videos are compressed and thus ignore the pre-computed information in compressed video streams. Motion vectors, one of the compression information, record the motion of the video frames. They can be directly extracted from the compression code stream without computational cost and serve as a solid prior for optical flow estimation. Therefore, we propose an optical flow model, MVFlow, which uses motion vectors to improve the speed and accuracy of optical flow estimation for compressed videos. In detail, MVFlow includes a key Motion-Vector Converting Module, which ensures that the motion vectors can be transformed into the same domain of optical flow and then be utilized fully by the flow estimation module. Meanwhile, we construct four optical flow datasets for compressed videos containing frames and motion vectors in pairs. The experimental results demonstrate the superiority of our proposed MVFlow, which can reduce the AEPE by 1.09 compared to existing models or save 52\% time to achieve similar accuracy to existing models.
\end{abstract}



\maketitle

\section{Introduction}
\label{sec:intro}

\begin{figure}
    \centering
    \includegraphics[width=\columnwidth]{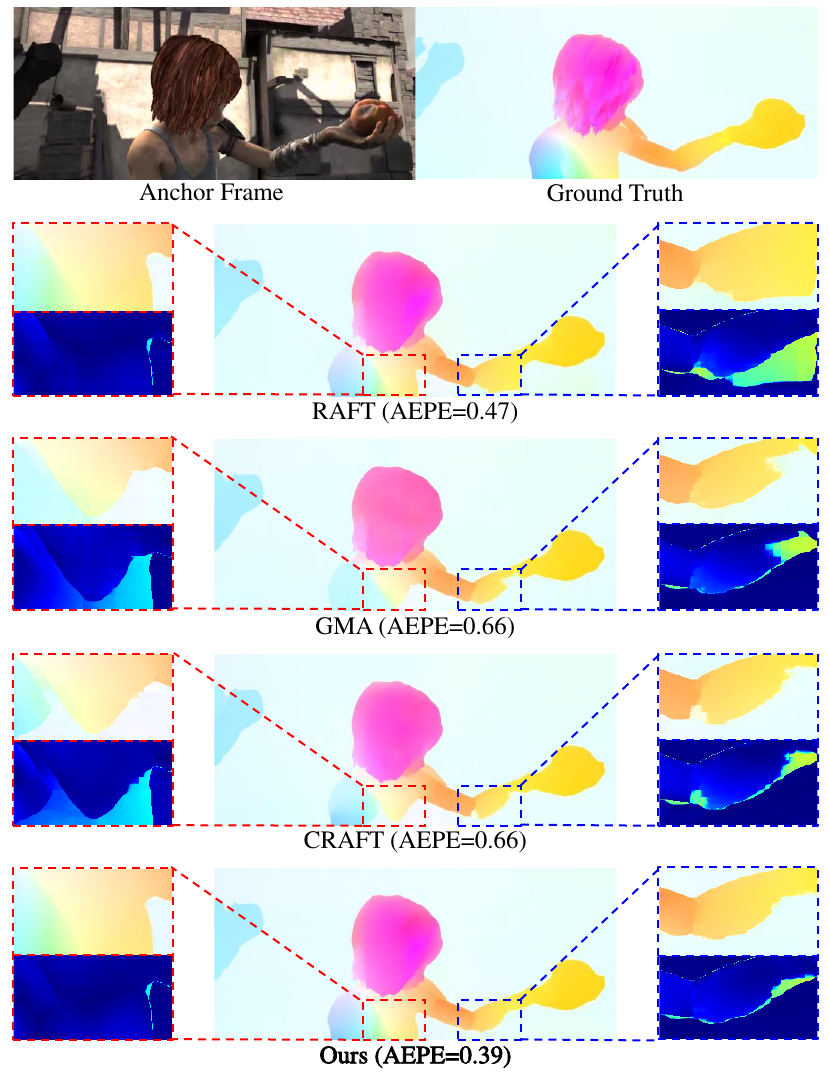}
    \caption{An example to show the superiority of our method. The estimated optical flow and the error map of different methods are visualized. Our method estimates less error and shows clear contours in the optical flow map.}
    \label{fig:head}
\end{figure}

Optical flow refers to the motion field between two frames, which is an important tool in computer vision and video processing. It has a wide range of application scenarios, including video super-resolution \cite{chan2022basicvsr++}, video frame interpolation \cite{huang2022real}, video inpainting \cite{xu2019deep}, object detection \cite{li2018flow} and tracking \cite{vihlman2020optical}, \textit{etc.} In recent years, with the development of deep learning and neural networks, many high-performance deep learning optical flow estimation models have emerged. These models learn discriminative features and then utilize the feature correlation between two frames to estimate accurate optical flow.
Although leaps and bounds have been made, the current optical flow estimation methods generally have a blind spot: they assume that the inputs are uncompressed high-quality frames, which is inconsistent with the practical situation. In fact, due to the huge amount of information, almost all videos are stored in a compressed format, which causes distortion and hinders the performance of existing optical flow models.

In order to find a solution, we need to look at the principles of video compression. The process of video compression can be divided into encoding and decoding. The basic idea of encoding is to dynamically divide the image into blocks and quantify them to discard the secondary information, thereby reducing the number of bits to store the video. Specifically, the compression algorithm also takes advantage of the video's temporal continuity by matching the blocks of adjacent frames and sharing the information between frames to reduce redundancy further. The matching offsets are called motion vectors and are stored together in the compressed video. When decoding, the algorithm reads the encoded blocks with the motion vectors to reconstruct the image for each frame.

The motion vectors have a close definition to optical flow and can be regarded as a rough block-level optical flow. Importantly, it is already pre-computed and can be extracted from the compression code stream without additional computational cost. A simple way to use motion vectors is to use them directly as the initial solution for iterative optical flow models such as RAFT \cite{teed2020raft}. However, we find that such an implementation fails to improve the accuracy of optical flow estimation. The main reason is that the existing deep optical flow models are better at handling smooth optical flow maps, while the sparse and block-level motion vectors do not fit this pattern and cannot be effectively utilized by those models. To exploit the motion vector prior, we propose our MVFlow with a Motion-Vector Converting Module (MVCM). The module can initially convert the domain of the motion vector map through the contextual correlation of the image, so that the motion information contained in motion vectors can be incorporated into the process of optical flow estimation. As shown in Figure \ref{fig:head}, our MVFlow demonstrates excellent performance and estimates accurate motion for the arm and body in the area marked by the boxes.

Besides, we construct the training and evaluation datasets for optical flow estimation of compressed videos to conduct our experiments. We compress the videos of four typical optical flow datasets (FlyingThings 3D \cite{mayer2016large}, MPI Sintel\cite{butler2012naturalistic} and KITTI 2012/2015 \cite{geiger2012we, menze2015object}) and extract the motion vectors from decoding.

In total, our contributions are:

\begin{itemize}
\item We propose a novel optical flow estimation framework that exploits video motion vectors as prior information for accurate and fast motion estimation for compressed videos. To the best of our knowledge, this is the first attempt that uses motion vectors to assist deep optical flow estimation.

\item To address the domain gap between motion vectors and optical flow, we propose a Motion-Vector Converting Module that utilizes the correlation of video content and motion to regulate motion vectors.

\item Experiments prove the superiority of MVFlow in terms of performance and efficiency. Compared to RAFT, MVFlow can reduce AEPE by 1.09 with the same iteration steps, or save 52\% computation time to reach similar accuracy.

\item For the first time, we construct four datasets containing optical flow, compressed frames and motion vectors of different compression qualities. We believe they can facilitate the research on optical flow estimation of compressed videos.

\end{itemize}

\begin{figure*}
    \centering
    \includegraphics[width=\textwidth]{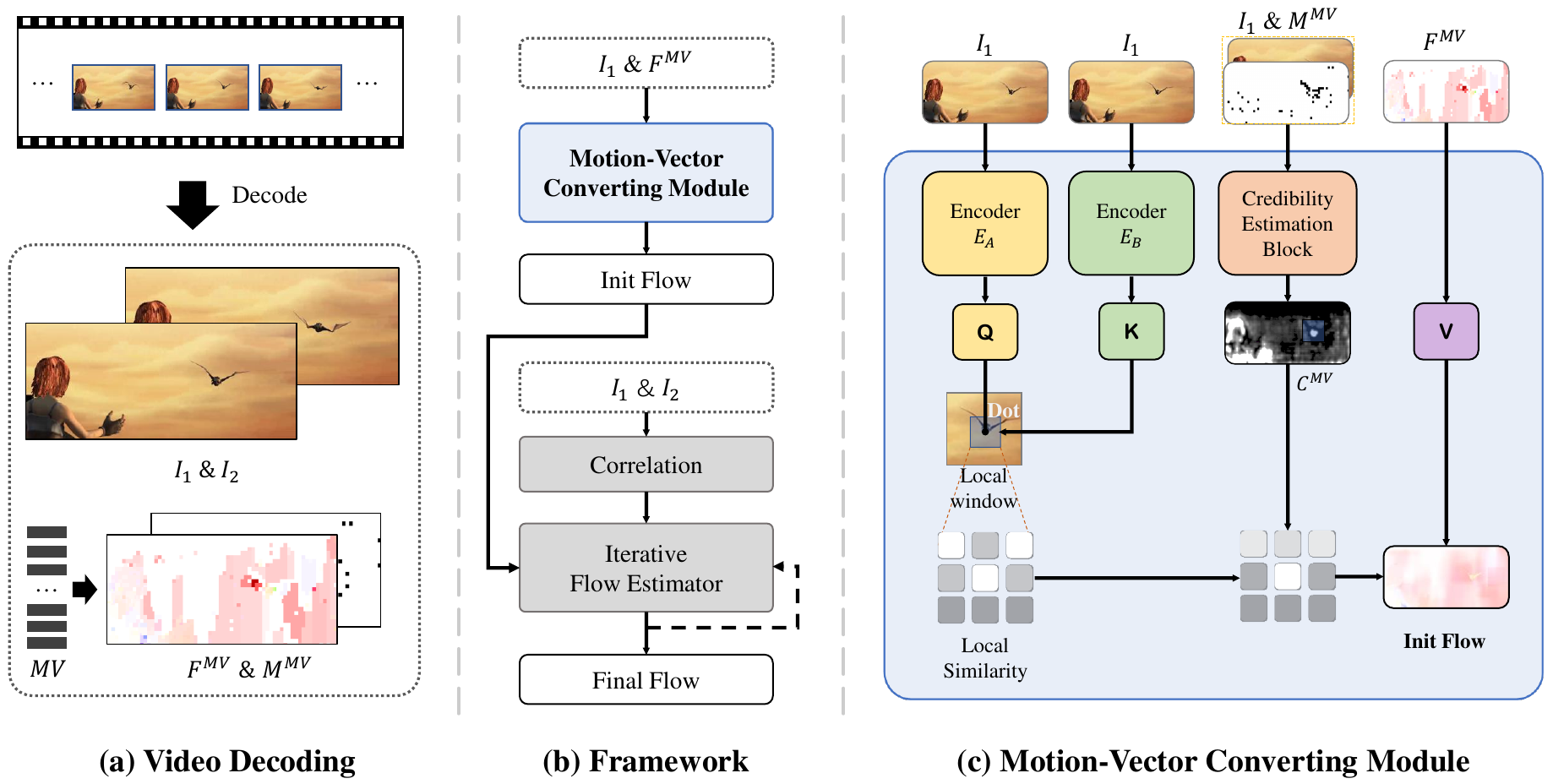}
    \caption{The architecture of our proposed MVFlow. (a) Decoding videos to get frames and the corresponding motion vectors. (b) The main framework of our MVFlow. (c) The structure of our Motion-Vector Converting Module.}
    \label{fig:main_method}
\end{figure*}
\section{Related Works}
\label{sec:related-work}
\subsection{Optical Flow}

Optical flow has been studied for a long time as a fundamental technology. Early on, according to the mathematical definition of optical flow, researchers design some traditional optical flow algorithms, such as Horn–Schunck \cite{horn1981determining} and Lucas-Kanade \cite{lucas1981iterative}. These methods can effectively estimate the optical flow of simple cases, but their accuracy is generally not good.

With the advent of deep learning, researchers have also begun to use deep neural networks for optical flow estimation. FlowNet \cite{dosovitskiy2015flownet} and FlowNet2.0 \cite{ilg2017flownet} are the first attempt that proves the feasibility of deep learning in optical flow estimation. After that, the multi-scale models \cite{sun2018pwc, hui2018liteflownet, hur2019iterative, zhao2020maskflownet} emerge. Next, RAFT \cite{teed2020raft} proposes an iterative method, which calculates global all-pair correlation and reuses it in every iteration. It has become the new baseline for subsequent researches. For example, various attention blocks \cite{sui2022craft, zhao2022gmflownet, Luo_2022_CVPR, huang2022flowformer} and big-kernel convolution layers \cite{sun2022skflow} are added to the components of RAFT to provide stronger representation and estimation capabilities. Meanwhile, global motion aggregation \cite{jiang2021learning} and global matching \cite{xu2022gmflow, zhao2022gmflownet} are also proposed to break the over-dependence on local cues of models.


Recently, some works have also begun to study optical flow estimation under different degradation conditions. For example, Zhang \emph{et al.}~\cite{OFID} gives a solution to estimate optical flow in the dark, and Argaw \emph{et al.}~\cite{argaw2021optical} tries to estimate optical flow from a single motion-blurred image. For compressed video, Young \emph{et al.}~\cite{young2020fast} introduces compression prior information into traditional variational optimization for optical flow estimation. However, it is not comparable to deep learning methods in terms of accuracy and speed. To the best of our knowledge, we are the first to exploit compressed priors in deep optical flow estimation.

\subsection{Video Compression}
Video compression has become an indispensable part of video processing, which can effectively save storage and transmission bandwidth. In recent years, some deep learning-based video compression algorithms~\cite{li2021deep,lu2019dvc,yang2020learning,hu2020improving} have been proposed with the expectation of achieving better compression performance. However, they are not currently available for practical applications due to the huge computational cost. Currently, commercial compression algorithms are still dominated by traditional methods\cite{wiegand2003overview,sullivan2012overview,bross2021developments}.

Inter-frame predictive coding is an important part of traditional video compression algorithms. It calculates the motion vectors to measure the motion information between frames and removes temporal redundancy based on them. Note that motion vectors can be extracted from the compressed video stream without additional computational cost at the receiver end. Recently, some works attempt to utilize motion vectors to assist various vision tasks~\cite{chen2021compressed,chen2020bitstream,xu2022accelerating,tan2020real,wu2018compressed,xu2016learning}. Chen \emph{et al.}~\cite{chen2020bitstream} first explore the compressed video super-resolution task, and improve the model performance by leveraging the interactivity between decoding prior and deep prior. Specifically, they align the features of different frames based on motion vectors. Similarly, Xu \emph{et al.}~\cite{xu2022accelerating} uses motion vectors to propagate segmentation masks from keyframes to other frames, which can improve the efficiency and performance of video object segmentation. Considering that the motion vectors represent the primary motion of the videos, we use them to improve the performance of optical flow estimation in our work.

\section{Proposed Method}
In this section, we first analyze our motivations and then provide an overview of our optical flow estimation framework with motion vector prior. Next, the structure of our proposed MVCM is described in detail. Finally, we extend our MVCM to incorporate the common warm-start strategy.
\subsection{Motivation}

Almost all videos exposed to non-professional users are stored in a compressed format. The mainstream video compression frameworks perform motion compensation between frames, so the compressed video stores a set of offsets to represent the motion between frames. Such offsets are called motion vectors, which can be obtained without extra computational cost.

Motion vectors and optical flow are both representations of motion between frames, but there are two differences. Firstly, motion vectors are block-level, while optical flow records pixel-level motion. Second, motion vectors are calculated locally during encoding. It differs from the estimated optical flow, which needs to find motion field from the context of the entire frame. Therefore, using the motion vectors as an additional input can help optical flow estimation from two perspectives: 1) The optical flow model can conduct iterative updates based on the rough solution given by motion vectors, making converging faster. 2) Due to the distortion caused by compression, the inter-frame correspondence for some regions is disrupted, so the optical flow models rely more on the learned global prior like smoothness and ignores some small objects that move independently. In contrast, motion vectors store the best matches found for each block individually, which can play an important complementary role in estimating optical flow of compressed video.

\begin{figure}
    \centering
    \includegraphics[width=\columnwidth]{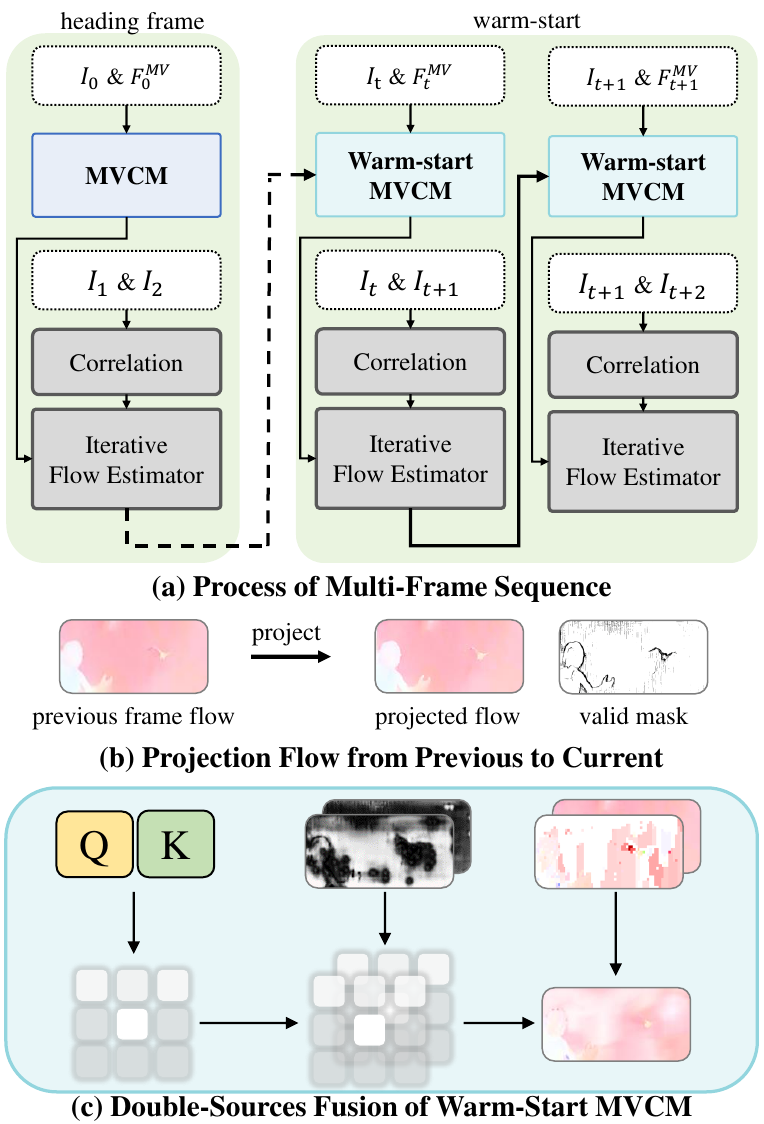}
    \caption{(a) Process of multi-frame sequence with warm-start strategy. For the heading frame, we use our ordinary MVCM, while for subsequent frames, we use warm-start MVCM, which can utilize the estimation of the previous frame. (b) Projection of flow from the previous frame to the current frame. (c) The modified aggregation process in warm-start MVCM.}
    \label{fig:warm-start}
\end{figure}
\subsection{Overview}
As shown in Figure \ref{fig:main_method}(a), we first decode the video to obtain consecutive frames and the corresponding motion vectors. We denote the first frame of the two frames as $I_1$, the second frame as $I_2$, and the motion vectors from the previous to the next frame as $MV$. The initial representation of $MV$ is a group of vectors, each of which records a compressed block's position, size, and motion offset. We convert the $MV$ into a dense flow map denoted as $F^{MV}$ by filling the pixels in each block with the same motion offset. Subsequently, we estimate the optical flow with $F^{MV}$ as additional input in our framework shown in Figure \ref{fig:main_method}(b). Our model is a variant based on RAFT \cite{teed2020raft}, called MVFlow. The estimation process of MVFlow contains three stages, of which the first two stages can be parallelized. In the first stage, we adopt our Motion-Vector Converting Module to convert $F^{MV}$ into a smoother coarse flow map according to the contextual information of $I_1$. In the second stage, we extract the features of $I_1$ and $I_2$ and calculate the correlation. In the last stage, we take the converted coarse flow as the initial value and refer to the correlation information to perform an iterative optimization process.
\begin{figure*}
    \centering
    \includegraphics[width=0.95\textwidth]{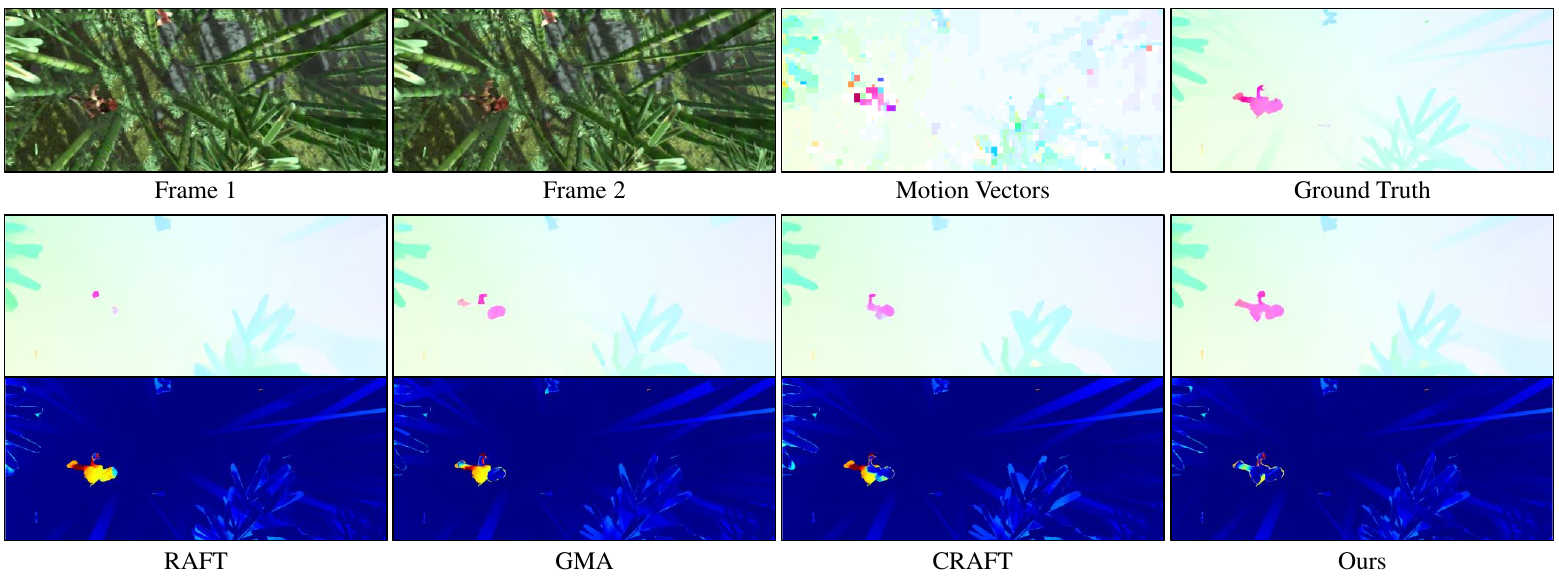}
    \caption{Qualitative comparison of our method and the state-of-the-art methods on Compressed Sintel. The displayed example is in QP of 37.}
    \label{fig:main_result}
\end{figure*}
\subsection{Motion-Vector Converting Module}
The $F^{MV}$ obtained directly from the motion vectors has a large domain gap with the optical flow, which can not be utilized effectively by existing deep learning-based optical flow estimation architectures, as proved in our experiment (Section \ref{sec:ablation}). Thus, we design a Motion-Vector Converting Module (MVCM) to convert $F^{MV}$ into the same domain of optical flow. Our inspiration consists of two parts. First, the $F^{MV}$ is sparse, and there are some regions without MV offsets, so we need to complement them with other regions. A good idea is to use the spatial correlation of $I_1$ to accomplish the filling process. Second, $F^{MV}$ also has some regions with inaccurate motion, which is caused by either the coarse block division or the matches not in line with actual motion. These wrong areas need to be figured out and corrected. We use the context information of $I_1$ to solve it. As a combination of the above two points, the specific design of our module is shown in Figure \ref{fig:main_method}(c), which follows the attention mechanism. As Equation \ref{eq:QKV}, $I_1$ is first fed into two different encoders to obtain Q and K maps, while $F^{MV}$ is directly taken as the V map, denoted as
\begin{equation}
    Q = E_A(I_1),~
    K = E_B(I_1),~
    V = F^{MV}
    \label{eq:QKV}
\end{equation}
$E_A$ and $E_B$ are two encoder blocks, each consisting of six convolution and corresponding activation layers. Then, in order to find the areas that need to be corrected, we use a Credibility Estimation Block to estimate the credibility of the motion prior for each pixel. This can be expressed as
\begin{equation}
    C^{MV} = CEB(I_1, {M}^{MV})
\end{equation}
where $M^{MV}$ refers to a mask indicating which regions have motion vectors, $C^{MV}$ is a weight map in the range $(0,1)$, and $CEB$ is a CNN block, which contains six convolution layers, six dilation convolution layers and their corresponding activation layers. The dilation convolution layers can extract broader contextual information to utilize spatial information comprehensively. The last activation function is sigmoid for limiting the range of $C^{MV}$.
We perform the correlation computation in local sliding windows instead of all pixels to avoid introducing too much extra computation. First, the correlation between the center pixel of the local window and other pixels is calculated:
\begin{equation}
    S_{i,j} = softmax(Q_{i,j} \cdot K_{i+k,j+l}),
\end{equation}
where $k,l \in [-d,d]$. The $\cdot$ mark refers to the vector dot product operator, so the computed correlation weight $S_{i,j}$ is a tensor of shape $(2d+1)^2$. Then the credibility of the pixels is combined with the correlation to get the final weights, donated as:
\begin{equation}
    \label{equ_weight_mv}
    W^{MV}_{i,j} = S_{i,j}\odot \hat{C}^{MV}_{i,j}
\end{equation}
The $\hat{C}^{MV}_{i,j}$ is a (2d+1)$\times$(2d+1) window extracted from $C^{MV}$ around pixel $(i,j)$. The $\odot$ mark refers to the element-wise product operator.
At last, we aggregate motions in local windows with the calculated weights:
\begin{equation}
    \label{equ_final}
    F_{i,j} = \frac{\sum_{k,l}{W^{MV}_{i,j,k,l} V_{i+k,j+l}}}{\sum_{k,l}{W^{MV}_{i,j,k,l}}}
\end{equation}

\subsection{Combination with Warm-Start Strategy}
\label{sec: ws}
In the practical setting for iterative optical flow estimation, a well-known strategy called warm-start uses the optical flow predicted for the previous frame as initialization, shown in Figure \ref{fig:warm-start}(a). Our method also provides an initialization from motion vectors. Therefore, to simultaneously utilize these two different sources of optical flow initialization, we design a warm-start MVCM module to fuse them. First, as shown in Figure \ref{fig:warm-start}(b), we need to project the flow of the previous frame to the current frame:
\begin{equation}
    F^{Prj} = FW(F^{Pre},F^{Pre}), ~
    M^{Prj} = FW(O, F^{Pre})
\end{equation}
$F^{Pre}$ is the flow estimation of the previous frame, and $F^{Prj}$ is the projected flow. $FW$ refers to forward warping, which will cause holes and overlaps. Thus, we calculate the valid mask $M^{Prj}$ of $F^{Prj}$ by forward warp an all-one matrix $O$.

Then, we send $M^{Prj}$, $M^{MV}$ and $I_1$ to a modified Credibility Estimation Block denoted as $CEB^{'}$, and get two credibility maps $C^{Prj}$ and $C^{MV}$, corresponding to $F^{Prj}$ and $F^{MV}$ respectively, denoted as
\begin{equation}
    C^{Prj},C^{MV} = CEB^{'}(I_1, M^{MV}, M^{Prj})
\end{equation}
Next, as shown in Figure \ref{fig:warm-start}(c), we calculate the weights of projected flow as Equation \ref{equ_weight_mv} and \ref{equ_weight_prj}.
\begin{equation}
    \label{equ_weight_prj}
    W^{Prj}_{i,j} = S_{i,j}\odot \hat{C}^{Prj}_{i,j}
\end{equation}
Finally, we replace Equation \ref{equ_final} with Equation \ref{equ_final_ws}, which means a fusion of the two different sources of initialization.
\begin{equation}
    \label{equ_final_ws}
    F_{i,j} = \frac{\sum_{k,l}{W^{MV}_{i,j,k,l} V^{MV}_{i+k,j+l} + W^{Prj}_{i,j,k,l} V^{Prj}_{i+k,j+l}}}{\sum_{k,l}{W^{MV}_{i,j,k,l} + W^{Prj}_{i,j,k,l}}},
\end{equation}
where $V^{MV}$ and $V^{Prj}$ correspond to $F^{MV}$ and $F^{Prj}$ respectively.

\begin{table*}[]
\center
\caption{Comparison of our method and three well-known optical flow methods. We adopt AEPE and F1 as the metrics for all datasets, and both are lower when the results are more accurate. For each QP setting, we color the best value for each column in red.}
\begin{tabular}{@{}clcccccccccccc@{}}
\toprule
\multirow{3}{*}{QP} & \multicolumn{1}{c}{\multirow{3}{*}{Method}} & \multicolumn{4}{c}{MPI Sintel}                                      & \multicolumn{4}{c}{KITTI 2012}                                      & \multicolumn{4}{c}{KITTI 2015}                                      \\ \cmidrule(l){3-14} 
                    & \multicolumn{1}{c}{}                        & \multicolumn{2}{c}{clean pass}   & \multicolumn{2}{c}{final pass}   & \multicolumn{2}{c}{NOC}          & \multicolumn{2}{c}{ALL}          & \multicolumn{2}{c}{NOC}          & \multicolumn{2}{c}{ALL}          \\ \cmidrule(l){3-14} 
                    & \multicolumn{1}{c}{}                        & AEPE          & F1               & AEPE          & F1               & AEPE          & F1               & AEPE          & F1               & AEPE          & F1               & AEPE          & F1               \\ \midrule
\multirow{6}{*}{22} & RAFT                                        & 1.90          & 6.09\%           & 3.48          & 11.58\%          & 1.34          & 7.01\%           & 2.78          & 13.50\%          & 3.13          & 12.96\%          & 6.54          & 21.19\%          \\
                    & GMA                                         & 2.07          & 6.11\%           & 4.43          & 13.37\%          & 1.45          & 7.85\%           & 2.68          & 14.09\%          & 3.45          & 14.30\%          & 6.74          & 21.92\%          \\
                    & CRAFT                                       & 1.95          & 7.41\%           & 3.95          & 13.32\%          & 1.50          & 8.27\%           & 2.79          & 14.67\%          & 3.60          & 14.94\%          & 6.59          & 22.66\%          \\
                    & GMFlow                                      & \textbf{1.66} & 5.75\%           & 3.86          & 12.59\%          & 1.79          & 9.29\%           & 3.47          & 15.81\%          & 3.31          & 15.68\%          & 6.91          & 23.19\%          \\
                    & GMFlowNet                                   & 2.13          & 8.29\%           & 4.22          & 14.71\%          & 1.34          & 6.69\%           & 2.68          & \textbf{12.57\%} & \textbf{2.85} & \textbf{11.68\%} & \textbf{5.97} & \textbf{19.26\%} \\
                    & Ours                                        & 1.85          & \textbf{5.56\%}  & \textbf{3.43} & \textbf{10.27\%} & \textbf{1.29} & \textbf{6.58\%}  & \textbf{2.59} & 12.60\%          & 3.13          & 12.80\%          & 6.07          & 20.64\%          \\ \midrule
\multirow{6}{*}{27} & RAFT                                        & 2.16          & 6.87\%           & 3.79          & 12.75\%          & 1.51          & 8.62\%           & 3.03          & 15.27\%          & 3.43          & 14.59\%          & 7.00          & 22.79\%          \\
                    & GMA                                         & 2.16          & 6.83\%           & \textbf{3.25} & \textbf{9.95\%}  & 1.76          & 9.80\%           & 3.12          & 16.05\%          & 3.89          & 15.87\%          & 7.41          & 23.40\%          \\
                    & CRAFT                                       & 2.07          & 8.06\%           & 4.20          & 14.40\%          & 1.73          & 10.24\%          & 3.15          & 16.60\%          & 3.84          & 16.64\%          & 7.29          & 24.36\%          \\
                    & GMFlow                                      & \textbf{1.89} & 6.48\%           & 4.09          & 13.92\%          & 2.03          & 11.01\%          & 3.84          & 17.71\%          & 3.77          & 17.54\%          & 7.66          & 24.91\%          \\
                    & GMFlowNet                                   & 2.48          & 9.43\%           & 4.46          & 16.15\%          & 1.59          & 8.49\%           & 3.11          & 14.54\%          & 3.34          & \textbf{13.37\%} & 6.79          & \textbf{20.84\%} \\
                    & Ours                                        & 2.01          & \textbf{6.20\%}  & 3.70          & 11.28\%          & \textbf{1.41} & \textbf{7.97\%}  & \textbf{2.83} & \textbf{14.20\%} & \textbf{3.17} & 13.98\%          & \textbf{6.25} & 21.82\%          \\ \midrule
\multirow{6}{*}{32} & RAFT                                        & 2.54          & 8.27\%           & 4.06          & 14.74\%          & 2.14          & 12.77\%          & 3.94          & 19.64\%          & 4.69          & 18.81\%          & 8.89          & 26.60\%          \\
                    & GMA                                         & 2.45          & 8.49\%           & 4.53          & 16.95\%          & 2.16          & 13.13\%          & 3.68          & 19.63\%          & 4.83          & 19.99\%          & 8.88          & 27.21\%          \\
                    & CRAFT                                       & 2.48          & 9.83\%           & 4.46          & 16.66\%          & 2.16          & 13.86\%          & 3.79          & 20.57\%          & 4.63          & 20.73\%          & 8.61          & 28.27\%          \\
                    & GMFlow                                      & \textbf{2.12} & 8.29\%           & 4.42          & 16.07\%          & 2.50          & 14.46\%          & 4.57          & 21.35\%          & 4.70          & 21.55\%          & 9.13          & 28.75\%          \\
                    & GMFlowNet                                   & 2.91          & 11.45\%          & 4.90          & 18.63\%          & 2.28          & 12.80\%          & 4.07          & 19.18\%          & 4.46          & 17.73\%          & 8.59          & \textbf{25.14\%} \\
                    & Ours                                        & 2.24          & \textbf{7.47\%}  & \textbf{4.01} & \textbf{13.22\%} & \textbf{1.88} & \textbf{11.58\%} & \textbf{3.48} & \textbf{18.19\%} & \textbf{4.09} & \textbf{17.62\%} & \textbf{7.66} & 25.30\%          \\ \midrule
\multirow{6}{*}{37} & RAFT                                        & 3.09          & 15.14\%          & 4.95          & 18.35\%          & 3.06          & 19.46\%          & 5.31          & 26.37\%          & 6.29          & 24.85\%          & 11.33         & 32.12\%          \\
                    & GMA                                         & 3.18          & 11.93\%          & 4.80          & 19.38\%          & 2.91          & 18.85\%          & \textbf{4.76} & 25.35\%          & 6.63          & 25.62\%          & 11.61         & 32.36\%          \\
                    & CRAFT                                       & 3.16          & 13.39\%          & 5.27          & 20.40\%          & 3.10          & 20.77\%          & 5.13          & 27.37\%          & 6.41          & 26.94\%          & 11.30         & 34.08\%          \\
                    & GMFlow                                      & \textbf{2.77} & 11.79\%          & 4.96          & 18.78\%          & 3.37          & 20.42\%          & 5.91          & 27.49\%          & 6.04          & 26.89\%          & 11.22         & 33.74\%          \\
                    & GMFlowNet                                   & 3.61          & 14.83\%          & 5.51          & 21.92\%          & 3.20          & 19.21\%          & 5.34          & 25.72\%          & 6.23          & 24.57\%          & 11.01         & 31.37\%          \\
                    & Ours                                        & 2.87          & \textbf{10.57\%} & \textbf{4.80} & \textbf{17.06\%} & \textbf{2.86} & \textbf{18.74\%} & 4.92          & \textbf{25.14\%} & \textbf{5.19} & \textbf{23.28\%} & \textbf{9.43} & \textbf{30.52\%} \\ \bottomrule
\end{tabular}
\label{tab:SOTA}
\end{table*}
\section{Experiments}

\subsection{Dataset Construction}
We make our compressed video optical flow dataset based on four existing datasets. They are: FlyingThings3D\cite{mayer2016large}, MPI Sintel(train) \cite{butler2012naturalistic}, KITTI 2012(train) \cite{geiger2012we} and KITTI 2015(train) \cite{menze2015object}. In our experiments, we use the H264 codec to compress the video because H264 is currently the most mature and widely used encoding tool. We first encode each sequence in the dataset with four different quantization parameters, 22, 27, 32, and 37. In order to maintain the consistency of the direction of motion vectors and optical flow, the videos are compressed in reverse order. Then, each frame and the corresponding motion vectors are decoded from the compressed videos. In the experiment, we use the Compressed FlyingThings3D as the training set, and the rest datasets are set as the evaluation benchmark. All the generated datasets will be uploaded to the public platform to facilitate future research.

\subsection{Settings}
We implement our model based on the code of RAFT\cite{teed2020raft}. The loss functions are added to all the intermediate flow estimations (including the output of MVCM) and trained the model for 120k steps on the aforementioned Compressed FlyingThings dataset. We use only four iterations in each step to speed up the training. We use AdamW \cite{adamw} optimizer and set weight\_decay=5e-5 and eps=1e-8. The learning rate is set to 1e-4 and decays linearly to 8.5e-5 during training. The batch\_size is set to 4. Our training device is a single Nvidia RTX 3090. For data augmentation, we randomly crop 800$\times$512 patches of the input frames for training. For better convergence, we use the original RAFT parameters on FlyingThings as the initialization parameters of those unmodified layers. At the same time, in order to train our model with the warm-start strategy, we fine-tune our model for an additional 30k steps, and the original MVCM parameters are fixed during fine-tuning. Other models for comparison that emerged in the experiments follow the same training process. Unless otherwise stated, all models are evaluated with 16 iterations.

\begin{table}[]
    \centering
    \caption{Ablation Study of four models, including the baseline model, the retrained model, the retrained model with MV inputs and our final model with MVCM. We also provide results on Compressed MPI Sintel and Compressed KITTI 2012 in the supplementary material.}
    \begin{tabular}{@{}clcccc@{}}
\toprule
\multirow{3}{*}{QP} & \multicolumn{1}{c}{\multirow{3}{*}{Method}} & \multicolumn{4}{c}{Compressed KITTI 2015}                           \\ \cmidrule(l){3-6} 
                    & \multicolumn{1}{c}{}                        & \multicolumn{2}{c}{NOC}          & \multicolumn{2}{c}{ALL}          \\ \cmidrule(l){3-6} 
                    & \multicolumn{1}{c}{}                        & AEPE          & F1               & AEPE          & F1               \\ \midrule
\multirow{4}{*}{22} & \textbf{Baseline}                           & 4.34          & 16.72\%          & 10.07         & 25.58\%          \\
                    & \textbf{$\uparrow$ + Retrain}               & \textbf{3.13} & 12.96\%          & 6.54          & 21.19\%          \\
                    & \textbf{$\uparrow$ + MV}                    & 3.48          & 13.39\%          & 6.98          & 21.61\%          \\
                    & \textbf{$\uparrow$ + MVCM}                  & \textbf{3.13} & \textbf{12.80\%} & \textbf{6.07} & \textbf{20.64\%} \\ \midrule
\multirow{4}{*}{27} & \textbf{Baseline}                           & 5.22          & 19.79\%          & 11.53         & 28.34\%          \\
                    & \textbf{$\uparrow$ + Retrain}               & 3.43          & 14.59\%          & 7.00          & 22.79\%          \\
                    & \textbf{$\uparrow$ + MV}                    & 3.63          & 14.93\%          & 7.36          & 23.07\%          \\
                    & \textbf{$\uparrow$ + MVCM}                  & \textbf{3.17} & \textbf{13.98\%} & \textbf{6.25} & \textbf{21.82\%} \\ \midrule
\multirow{4}{*}{32} & \textbf{Baseline}                           & 7.36          & 27.13\%          & 14.53         & 34.74\%          \\
                    & \textbf{$\uparrow$ + Retrain}               & 4.69          & 18.81\%          & 8.89          & 26.60\%          \\
                    & \textbf{$\uparrow$ + MV}                    & 4.90          & 19.27\%          & 9.27          & 27.02\%          \\
                    & \textbf{$\uparrow$ + MVCM}                  & \textbf{4.09} & \textbf{17.62\%} & \textbf{7.66} & \textbf{25.30\%} \\ \midrule
\multirow{4}{*}{37} & \textbf{Baseline}                           & 9.67          & 35.58\%          & 17.52         & 42.11\%          \\
                    & \textbf{$\uparrow$ + Retrain}               & 6.29          & 24.85\%          & 11.33         & 32.12\%          \\
                    & \textbf{$\uparrow$ + MV}                    & 6.55          & 25.38\%          & 11.80         & 32.57\%          \\
                    & \textbf{$\uparrow$ + MVCM}                  & \textbf{5.19} & \textbf{23.28\%} & \textbf{9.43} & \textbf{30.52\%} \\ \bottomrule
\end{tabular}
    \label{tab:ablation}
\end{table}
\subsection{Comparison with the State-of-the-Art Methods}
We first compare our MVFlow with five well-known optical flow methods. They are RAFT\cite{teed2020raft}, GMA\cite{jiang2021learning}, CRAFT\cite{sui2022craft}, GMFlow\cite{xu2022gmflow} (GMF) and GMFlowNet\cite{zhao2022gmflownet} (GMFNet). Because off-the-shelf optical flow models do not perform well on compressed video (as shown in Supplementary Materials), all models are retrained with the same settings as ours. AEPE (Average Endpoint Error) and F1 (percentage of outliers) are chosen as metrics in our experiment. The results are shown in Tab \ref{tab:SOTA}. As we can see, in the vast majority of comparisons, our method shows clear superiority. 

An interesting pattern is that although the performance of RAFT, GMA, and CRAFT is progressively improved on the uncompressed optical flow test set, CRAFT and GMA do not outperform RAFT in compressed videos. This may be due to the lack of flexibility caused by the large amount of attention computation introduced by GMA and CRAFT.

We can also find that our method leads by a more significant margin at a higher QP. The reason is that higher QP introduces more compression noise, making motion estimation more challenging. Despite retraining on Compressed FlyingThings, RAFT, GMA and CRAFT still fail to find correct motion from the compressed videos. Unlike them, our method can handle this situation by exploiting the motion vectors.

We also give an example for qualitative comparison in Figure \ref{fig:main_result}, which are from Compressed Sintel dataset. The methods without utilizing motion vectors fail to estimate the flow of the human in the first example and the window in the second example. With motion vectors as additional hints, our method generates finer initializations, thus better handling these complex cases. The example from KITTI 2015 dataset can be found in the appendix.
\begin{figure}
    \centering
    \includegraphics[width=\columnwidth]{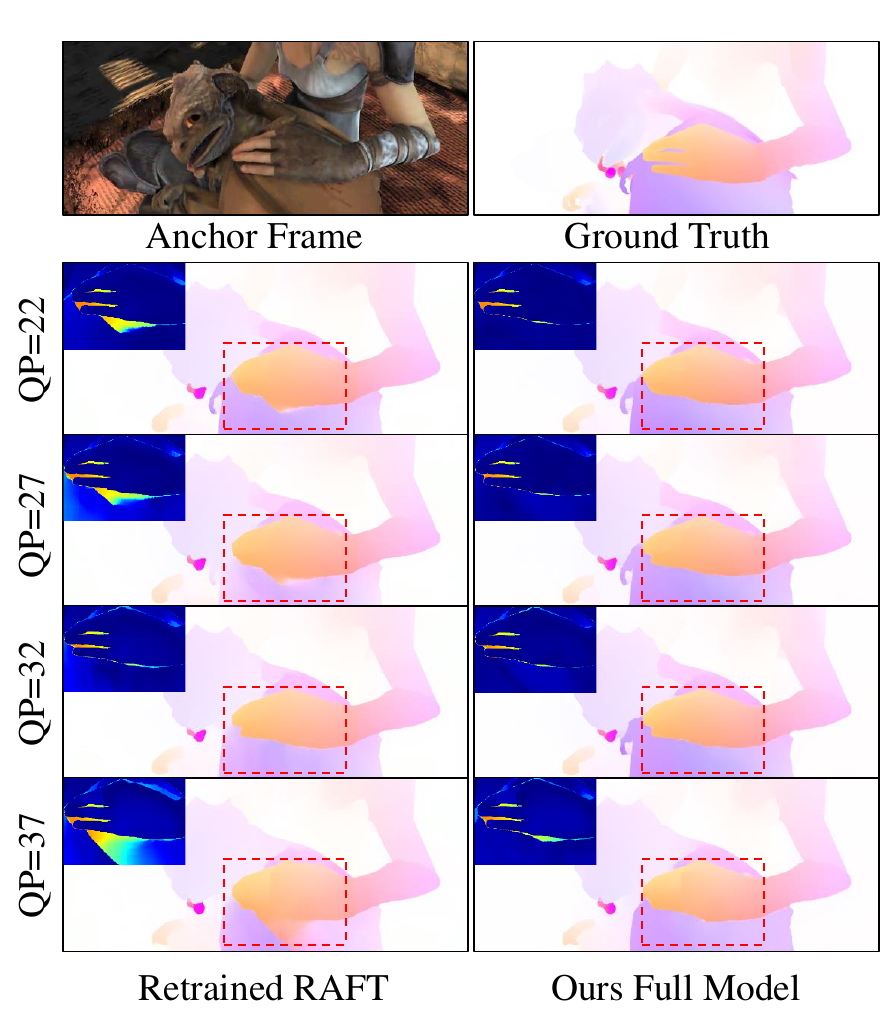}
    \caption{The comparison of our MVFlow and the retrained RAFT on a group of frames with different QP. RAFT produces various wrong estimates due to information loss during video compression, while our model, while our model shows robustness on different QP.}
    \label{fig:QP}
\end{figure}
\begin{table}[]
    \centering
    \caption{Comparison of different initialization strategies. }
    \begin{tabular}{@{}clcccc@{}}
\toprule
\multirow{3}{*}{QP} & \multicolumn{1}{c}{\multirow{3}{*}{Method}} & \multicolumn{4}{c}{Compressed MPI Sintel}                           \\ \cmidrule(l){3-6} 
                    & \multicolumn{1}{c}{}                        & \multicolumn{2}{c}{clean pass}   & \multicolumn{2}{c}{final pass}   \\ \cmidrule(l){3-6} 
                    & \multicolumn{1}{c}{}                        & AEPE          & F1               & AEPE          & F1               \\ \midrule
\multirow{4}{*}{22} & Zero                                        & 1.90          & 6.09\%           & 3.48          & 11.58\%          \\
                    & Warm-Start                                  & 1.83          & 6.12\%           & 3.46          & 11.37\%          \\
                    & \textbf{MVCM}                               & 1.85          & \textbf{5.56\%}  & 3.43          & \textbf{10.27\%} \\
                    & \textbf{MVCM + Warm-Start}                  & \textbf{1.71} & 5.71\%           & \textbf{3.28} & 10.64\%          \\ \midrule
\multirow{4}{*}{27} & Zero                                        & 2.16          & 6.87\%           & 3.79          & 12.75\%          \\
                    & Warm-Start                                  & 2.01          & 6.86\%           & 3.58          & 12.58\%          \\
                    & \textbf{MVCM}                               & 2.01          & \textbf{6.20\%}  & 3.70          & \textbf{11.28\%} \\
                    & \textbf{MVCM + Warm-Start}                  & \textbf{1.80} & 6.40\%           & \textbf{3.50} & 11.74\%          \\ \midrule
\multirow{4}{*}{32} & Zero                                        & 2.54          & 8.27\%           & 4.06          & 14.74\%          \\
                    & Warm-Start                                  & 2.33          & 8.30\%           & 4.05          & 14.78\%          \\
                    & \textbf{MVCM}                               & 2.24          & \textbf{7.47\%}  & 4.01          & \textbf{13.22\%} \\
                    & \textbf{MVCM + Warm-Start}                  & \textbf{2.11} & 7.74\%           & \textbf{3.94} & 13.88\%          \\ \midrule
\multirow{4}{*}{37} & Zero                                        & 3.09          & 15.14\%          & 4.95          & 18.35\%          \\
                    & Warm-Start                                  & 3.17          & 11.67\%          & 4.91          & 18.46\%          \\
                    & \textbf{MVCM}                               & 2.87          & \textbf{10.57\%} & 4.80          & \textbf{17.06\%} \\
                    & \textbf{MVCM + Warm-Start}                  & \textbf{2.86} & 10.71\%          & \textbf{4.43} & 17.39\%          \\ \bottomrule
\end{tabular}
    \label{tab:ws}
\end{table}
\begin{figure*}
    \centering
    \includegraphics[width=0.95\textwidth]{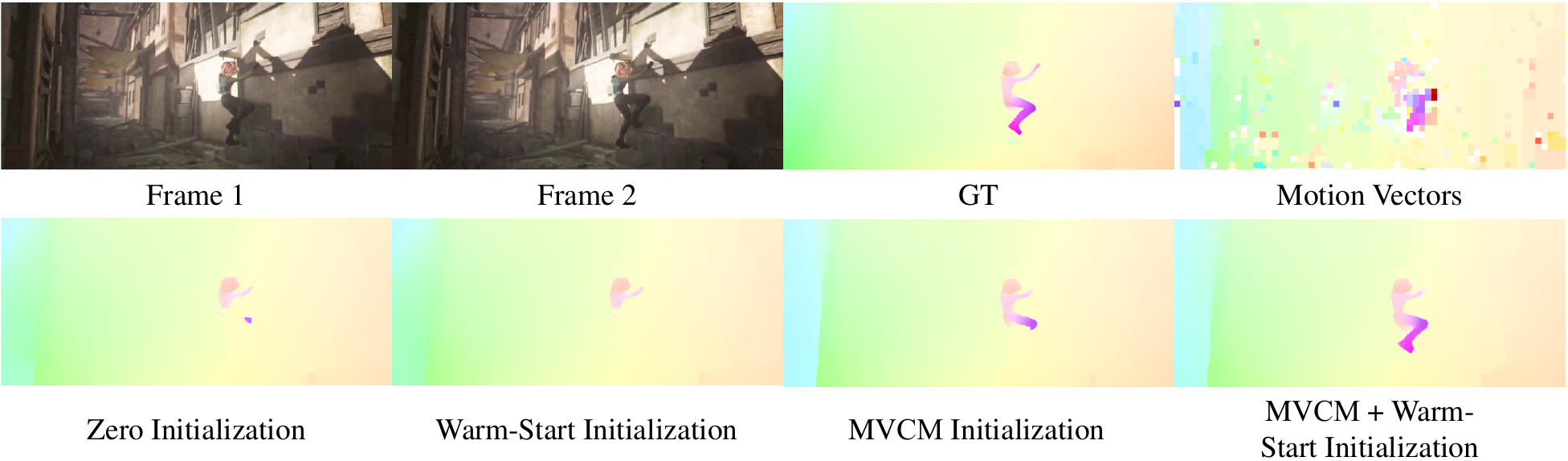}
    \caption{A group of qualitative examples of different initialization methods.}
    \label{fig:strategy}
\end{figure*}
\subsection{Ablation Study}
\label{sec:ablation}
We design a set of ablation experiments to probe the effect of each of our modifications. A total of four models are compared in the experiment, and the results can be found in Table \ref{tab:ablation}. The first model is the baseline, which directly uses the pre-trained parameters of RAFT (raft-things.pth). The second model is retrained on our Compressed FlyingThings and is thus more robust to compression noise. The third model simply adds motion vectors for initialization based on the second model. Experiments show that this naive scheme brings negative lift. The last model, our full MVFlow, adds MVCM as a preprocessing module, which converts the motion vectors to the same domain of optical flow. We can see in the table that MVCM brings a significant improvement. To show the effectiveness and robustness of our proposed method, we give qualitative comparisons on different QP settings in Figure \ref{fig:QP}.
\begin{figure}
    \centering
    \includegraphics[width=\columnwidth]{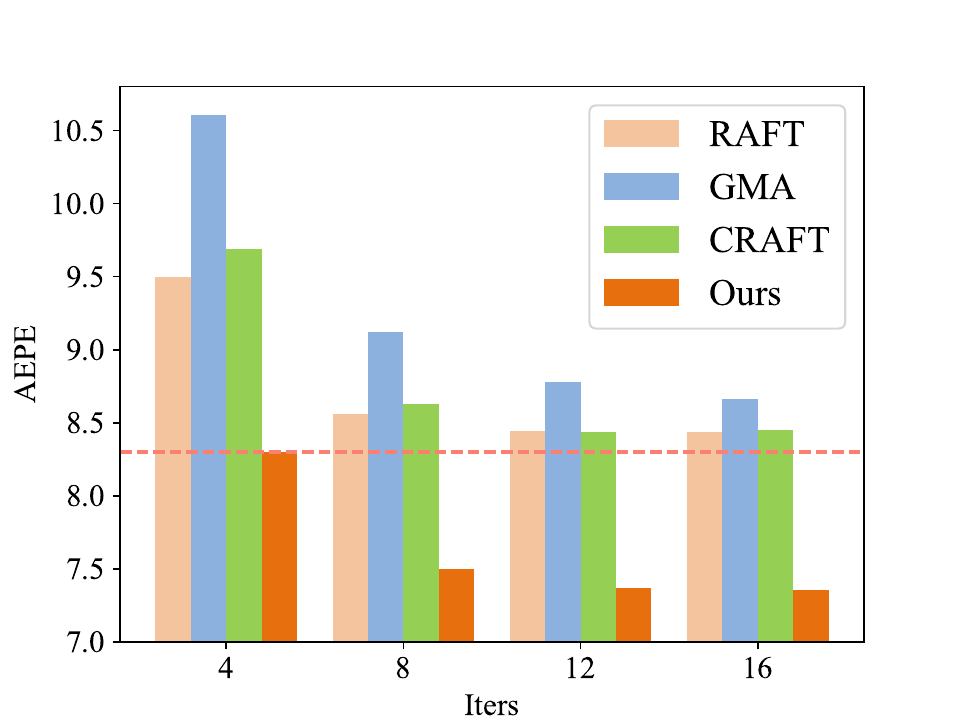}
    \caption{Efficiency comparison of four methods on Compressed KITTI 2015 dataset. The red dashed line highlights the performance of our method with four iterations}
    \label{fig:iters}
\end{figure}
\subsection{Discussion}
\textbf{Warm-Start Strategy} As mentioned in Section \ref{sec: ws}, warm-start is a common strategy used in iterative optical flow estimation methods for videos. It has a commonality with our method, that is, both of them give an initialized flow estimation for iteration. We evaluate four settings on Compressed MPI Sintel dataset to compare different initialization methods. They are the model with zero initialization, the model with warm-start initialization, the model with our MV initialization, and the model with combined strategy introduced in Section \ref{sec: ws}. The results are shown in Table \ref{tab:ws}, from where we can find that our combined strategy gets the best AEPE score, and our initialization gets the best F1 score. This means that the combined strategy brings an overall improvement compared to only using motion vectors, but the robustness to some problematic areas is reduced. Overall, both of our initialization strategies outperform the simple warm-start strategy. Figure \ref{fig:strategy} gives qualitative examples of different initialization methods. The motion vectors provide clear guidelines for the movement of the character's leg, thus enabling fine-grained optical flow estimation.

\textbf{Computational Efficiency} We compare the accuracy of different models with different iteration steps and give the result in Figure \ref{fig:iters} and Table \ref{tab:iters}. It can be clearly seen that our model only needs four iterations to outperform the results of other models with 16 iterations. This means that our model has a vast efficiency advantage under the requirement of achieving the same accuracy. Specifically, on an Nvidia RTX 3090, RAFT takes an average of 91ms to perform 16 iterations. In comparison, our method only needs four iterations that take 44ms to achieve comparable results, saving 52\% of computation time, which brings many benefits for practical use. On the other hand, our method outperforms RAFT by 1.09 of AEPE under the same iteration steps with only a slight increase in runtime.

\begin{table}[]
    \centering
    \setlength\tabcolsep{2pt}
    \caption{Running time required to achieve similar accuracy. Meanwhile, the performance under the same iterations is shown in the last column.}
\begin{tabular}{@{}lccccccc@{}}
\toprule
                 & RAFT  & GMA   & CRAFT  & GMF & GMFNet & \textbf{Ours}  & \textbf{(Ours)} \\ \midrule
Iterations       & 16    & 16    & 16     & -      & 16        & 4              & 16              \\
Runtime      & 91ms  & 119ms & 362ms  & 125ms  & 174ms     & \textbf{44ms}  & 99ms            \\
$\Delta$ Runtime & -0\%  & +31\% & +298\% & +37\%  & +91\%     & \textbf{-52\%} & +9\%            \\
AEPE             & 8.44  & 8.66  & 8.45   & 8.73   & \textbf{8.09}      & 8.30  & \textbf{7.35}   \\
$\Delta$ AEPE    & -0.00 & +0.22 & +0.01  & +0.29  & \textbf{-0.35}     & -0.14 & \textbf{-1.09}  \\ \bottomrule
\end{tabular}
    \label{tab:iters}
\end{table}

\section{Conclusion}
Optical flow estimation is an essential technique in the field of computer vision and video processing. However, almost all the videos are compressed. Existing methods ignore the powerful compression prior, thus fail to handle frames with compression noise. In this paper, we introduce the motion vectors in the compressed video stream to optical flow estimation. Our proposed MVFlow contains a Motion-Vector Converting Module to convert the motion vectors to the same domain of optical flow to better estimate the optical flow. We also construct four optical flow datasets for compressed videos. The experiments show that our proposed method is superior in effectiveness and efficiency.

\bibliographystyle{ACM-Reference-Format}
\bibliography{mybib}

\appendix
\section{Qualitative comparison on KITTI 2015}
\begin{figure*}
\includegraphics[width=\textwidth]{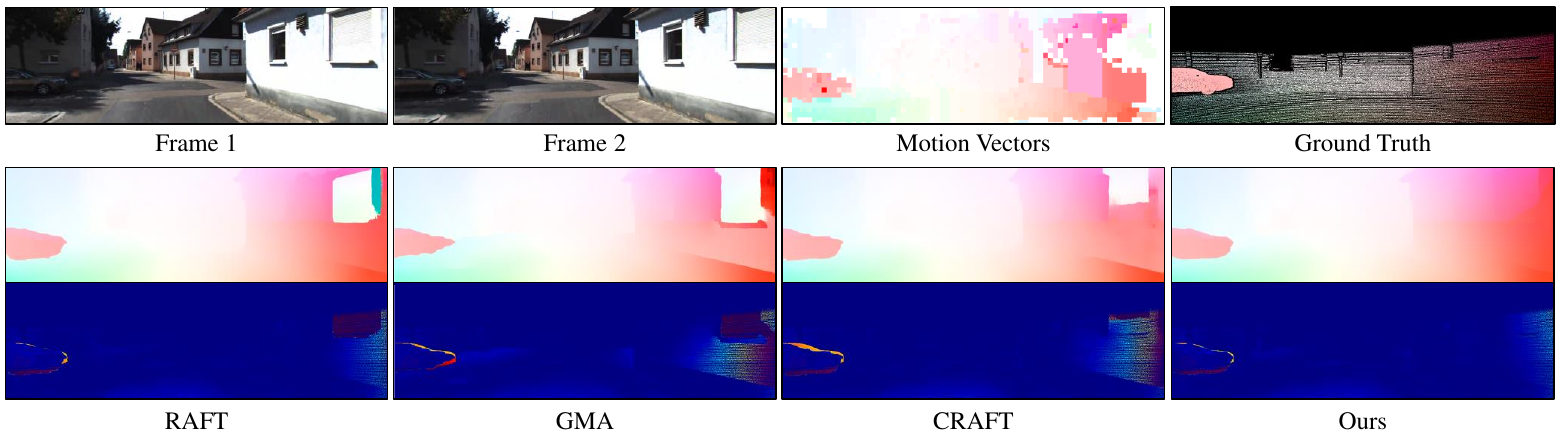}
\caption{Qualitative comparison on Compressed KITTI 2015.}
\label{fig:another_half}
\end{figure*}

We supplement a set of visual comparisons on Compressed KITTI 2015 in Figure \ref{fig:another_half}, where our method estimates a more complete optical flow map.

\subsection{The Necessity of Retraining the State-of-the-art Models}
As mentioned in the main manuscript, the off-the-shelf optical flow estimation models are not trained with compressed videos. Thus it cannot handle the compression noise well. For a fair comparison, we need to fine-tune the state-of-the-art model using the compressed data and settings the same as Ours. Table \ref{tab:retrain} shows the comparisons of RAFT \cite{teed2020raft}, GMA\cite{jiang2021learning}, CRAFT\cite{sui2022craft}, GMFlow\cite{xu2022gmflow} and GMFlowNet\cite{zhao2022gmflownet}.
The fine-tuning improves the performance of these methods for optical flow estimation on compressed videos, removing the influence of different training data and strengthening our experiments' rigor. The only exception is that GMFlowNet's performance on Compressed MPI Sintel decreased slightly after fine-tune, which may be due to the complex POLA structure and the Global Matching operation of GMFlowNet are sensitive to the distribution difference between Compressed Things and compressed MPI Sintel. However, GMFlowNet shows a very significant performance improvement on Compressed KITTI 2015 after fine-tuning, which still proves the role of retraining.

\begin{table}[]
\small
\setlength\tabcolsep{2pt}
\caption{The performance of state-of-the-art methods before and after fine-tuning. We test on two datasets including Compreesed MPI Sintel and Compressed KITTI 2015. The values given are the average test results on all QPs (22, 27,32,37).}
\begin{tabular}{@{}lcccccccc@{}}
\toprule
\multicolumn{1}{c}{\multirow{3}{*}{Method}} & \multicolumn{4}{c}{Compressed MPI Sintel}                          & \multicolumn{4}{c}{Compressed KITTI 2015}                           \\ \cmidrule(l){2-9} 
\multicolumn{1}{c}{}                        & \multicolumn{2}{c}{clean pass}  & \multicolumn{2}{c}{final pass}   & \multicolumn{2}{c}{NOC}          & \multicolumn{2}{c}{ALL}          \\ \cmidrule(l){2-9} 
\multicolumn{1}{c}{}                        & AEPE          & F1              & AEPE          & F1               & AEPE          & F1               & AEPE          & F1               \\ \midrule
RAFT                                        & 2.98          & 10.51\%         & 5.09          & 18.14\%          & 6.65          & 24.81\%          & 10.07         & \textbf{25.58\%} \\
RAFT-ft                                     & \textbf{2.42} & \textbf{8.19\%} & \textbf{4.07} & \textbf{14.36\%} & \textbf{4.38} & \textbf{17.80\%} & \textbf{8.44} & 25.67\%          \\ \midrule
GMA                                        & 2.52          & 9.68\%          & 4.58          & 17.87\%          & 5.48          & 21.00\%          & 9.95          & 27.96\%          \\
GMA-ft                                      & \textbf{2.46} & \textbf{8.34\%} & \textbf{4.09} & \textbf{14.80\%} & \textbf{4.70} & \textbf{18.95\%} & \textbf{8.66} & \textbf{26.22\%} \\ \midrule
CRAFT                                      & 2.39          & 9.43\%          & 4.42          & 17.34\%          & 5.56          & 20.89\%          & 9.91          & 27.70\%          \\
CRAFT-ft                                    & \textbf{2.10} & \textbf{8.08\%} & \textbf{3.86} & \textbf{11.33\%} & \textbf{4.62} & \textbf{18.62\%} & \textbf{8.55} & \textbf{25.94\%} \\ \midrule
GMFlow                                     & 2.64          & 10.58\%          & 4.94          & 19.24\%          & 5.93          & 24.53\%          & 11.45          & 31.49\%          \\
GMFlow-ft                                    & \textbf{2.11} & \textbf{8.08\%} & \textbf{4.33} & \textbf{15.34\%} & \textbf{4.46} & \textbf{20.41\%} & \textbf{8.73} & \textbf{27.65\%} \\ \midrule
GMFlowNet                                      & \textbf{2.48}          & \textbf{9.27\%}          & \textbf{4.72}          & \textbf{16.82\%}          & 5.58          & 20.97\%          & 10.29          & 27.68\%          \\
GMFlowNet-ft                                    & 2.78 & 11.00\% & 4.78 & 17.85\% & \textbf{4.22} & \textbf{16.84\%} & \textbf{8.09} & \textbf{24.15\%}
\\ \bottomrule
\end{tabular}
\label{tab:retrain}
\end{table}

\section{Full Ablation Study on Three Datasets}
Due to the limited number of pages, we only give the results of the ablation experiment on Compressed KITTI 2015 in the main manuscript. Here, we present the complete ablation experiments on three datasets in Table \ref{tab:ablation}. Similar to the results on Compressed KITTI 2015, the retraining and our proposed MVCM bring significant improvements. Using MV directly as initialization brings slight improvement on Compressed MPI Sintel, and even slightly hurts the performance on Compressed KITTI 2012/2015, further proving the necessity and effectiveness of our proposed MVCM.

\begin{table*}
    \centering
    \caption{Complete ablation study on three datasets.}
    \begin{tabular}{@{}clcccccccccccc@{}}
\toprule
\multirow{3}{*}{QP} & \multicolumn{1}{c}{\multirow{3}{*}{Method}} & \multicolumn{4}{c}{Compressed MPI Sintel}                                      & \multicolumn{4}{c}{Compressed KITTI 2012}                                      & \multicolumn{4}{c}{Compressed KITTI 2015}                                      \\ \cmidrule(l){3-14} 
                    & \multicolumn{1}{c}{}                        & \multicolumn{2}{c}{clean pass}   & \multicolumn{2}{c}{final pass}   & \multicolumn{2}{c}{NOC}          & \multicolumn{2}{c}{ALL}          & \multicolumn{2}{c}{NOC}          & \multicolumn{2}{c}{ALL}          \\ \cmidrule(l){3-14} 
                    & \multicolumn{1}{c}{}                        & AEPE          & F1               & AEPE          & F1               & AEPE          & F1               & AEPE          & F1               & AEPE          & F1               & AEPE          & F1               \\ \midrule
\multirow{4}{*}{22} & Baseline                                    & 2.07          & 6.19\%           & 3.99          & 12.62\%          & 1.93          & 9.21\%           & 4.98          & 18.18\%          & 4.34          & 16.72\%          & 10.07         & 25.58\%          \\
                    & $\uparrow$ + Retrain                        & 1.90          & 6.09\%           & 3.48          & 11.58\%          & 1.34          & 7.01\%           & 2.78          & 13.50\%          & \textbf{3.13} & 12.96\%          & 6.54          & 21.19\%          \\
                    & $\uparrow$ + MV                             & 1.90          & 6.04\%           & 3.48          & 11.14\%          & 1.36          & 7.03\%           & 2.82          & 13.52\%          & 3.48          & 13.39\%          & 6.98          & 21.61\%          \\
                    & $\uparrow$ + MVCM                           & \textbf{1.85} & \textbf{5.56\%}  & \textbf{3.43} & \textbf{10.27\%} & \textbf{1.29} & \textbf{6.58\%}  & \textbf{2.59} & \textbf{12.60\%} & \textbf{3.13} & \textbf{12.80\%} & \textbf{6.07} & \textbf{20.64\%} \\ \midrule
\multirow{4}{*}{27} & Baseline                                    & 2.40          & 7.76\%           & 4.44          & 15.42\%          & 2.52          & 14.04\%          & 5.85          & 22.70\%          & 5.22          & 19.79\%          & 11.53         & 28.34\%          \\
                    & $\uparrow$ + Retrain                        & 2.16          & 6.87\%           & 3.79          & 12.75\%          & 1.51          & 8.62\%           & 3.03          & 15.27\%          & 3.43          & 14.59\%          & 7.00          & 22.79\%          \\
                    & $\uparrow$ + MV                             & 2.13          & 6.82\%           & 3.77          & 12.38\%          & 1.57          & 8.76\%           & 3.14          & 15.40\%          & 3.63          & 14.93\%          & 7.36          & 23.07\%          \\
                    & $\uparrow$ + MVCM                           & \textbf{2.01} & \textbf{6.20\%}  & \textbf{3.70} & \textbf{11.28\%} & \textbf{1.41} & \textbf{7.97\%}  & \textbf{2.83} & \textbf{14.20\%} & \textbf{3.17} & \textbf{13.98\%} & \textbf{6.25} & \textbf{21.82\%} \\ \midrule
\multirow{4}{*}{32} & Baseline                                    & 3.02          & 11.10\%          & 5.32          & 19.54\%          & 3.94          & 21.79\%          & 7.83          & 29.86\%          & 7.36          & 27.13\%          & 14.53         & 34.74\%          \\
                    & $\uparrow$ + Retrain                        & 2.54          & 8.27\%           & 4.06          & 14.74\%          & 2.14          & 12.77\%          & 3.94          & 19.64\%          & 4.69          & 18.81\%          & 8.89          & 26.60\%          \\
                    & $\uparrow$ + MV                             & 2.43          & 8.26\%           & 4.10          & 14.71\%          & 2.18          & 12.91\%          & 4.03          & 19.78\%          & 4.90          & 19.27\%          & 9.27          & 27.02\%          \\
                    & $\uparrow$ + MVCM                           & \textbf{2.24} & \textbf{7.47\%}  & \textbf{4.01} & \textbf{13.22\%} & \textbf{1.88} & \textbf{11.58\%} & \textbf{3.48} & \textbf{18.19\%} & \textbf{4.09} & \textbf{17.62\%} & \textbf{7.66} & \textbf{25.30\%} \\ \midrule
\multirow{4}{*}{37} & Baseline                                    & 4.44          & 16.97\%          & 6.59          & 25.00\%          & 5.47          & 32.69\%          & 9.88          & 39.93\%          & 9.67          & 35.58\%          & 17.52         & 42.11\%          \\
                    & $\uparrow$ + Retrain                        & 3.09          & 15.14\%          & 4.95          & 18.35\%          & 3.06          & 19.46\%          & 5.31          & 26.37\%          & 6.29          & 24.85\%          & 11.33         & 32.12\%          \\
                    & $\uparrow$ + MV                             & 3.10          & 11.51\%          & 5.00          & 18.55\%          & 3.26          & 19.95\%          & 5.67          & 26.81\%          & 6.55          & 25.38\%          & 11.80         & 32.57\%          \\
                    & $\uparrow$ + MVCM                           & \textbf{2.87} & \textbf{10.57\%} & \textbf{4.80} & \textbf{17.06\%} & \textbf{2.86} & \textbf{18.74\%} & \textbf{4.92} & \textbf{25.14\%} & \textbf{5.19} & \textbf{23.28\%} & \textbf{9.43} & \textbf{30.52\%} \\ \cmidrule(l){2-14} 
\end{tabular}
\label{tab:ablation}
\end{table*}

\section{More Discussion}
\subsection{Validation on Clean Frames} For direct comparison with off-the-shelf optical flow models and further verifying the effect of our MVCM, we also test the performance of our model on clean frames. Our settings are consistent with RAFT, and the motion vectors under QP 22 (with the highest quality) are added as the input of MVCM. As shown in Table \ref{tab:clean}, our model can outperform the baseline RAFT on both the Sintel and KITTI datasets. At the same time, we also submitted our test results on Sintel and KTTI benchmarks and achieved performance beyond RAFT.

\subsection{Assisting Downstream Tasks} Accurate alignment is critical in video processing tasks. We design an experiment to demonstrate that our method can provide better alignments for downstream tasks. The experiment uses a composite task: given two compressed frames, use a U-Net model to synthesize the denoised two frames and the intermediate frame. The input of U-Net contains the original and the warped frames. In our comparison, the U-Net structure remains unchanged, and we only replace the optical flow used in warping. We choose QP 37 in this experiment because it is the most common in practice. The results are shown in Table \ref{tab:downstream}, which shows that the optical flow estimated by our method is more suitable for assisting compression video-related tasks.

\subsection{Validation on HEVC Codec}
To verify the flexibility of our method, we add an experiment with HEVC(H.265) codec. We adopt HEVC official lowdelay P coding settings, and set the reference frame to the previous frame (-1). We conduct the experiments on the clean pass of Sintel datasets 
with QP 32. As shown in Table \ref{tab:hevc}, our method is able to work well for other codecs, which demonstrates the generalization performance of our method.

\begin{table}[]
\centering
\caption{Comparison on clean optical flow datasets.}
\setlength\tabcolsep{3pt}
\begin{tabular}{@{}lccccccc@{}}
\toprule
Train Data & \multicolumn{4}{c}{C+T}                                          & \multicolumn{3}{c}{C+T+S+K(+H)}                  \\ \midrule
Method        & \multicolumn{2}{c}{Sintel(val)} & \multicolumn{2}{c}{KITTI(val)} & \multicolumn{2}{c}{Sintel(test)} & KITTI(test)   \\ \midrule
RAFT          & 1.43           & 2.71           & 5.04          & 17.40          & 1.61            & 2.86           & 5.10          \\
Ours          & \textbf{1.38}  & \textbf{2.67}  & \textbf{4.66} & \textbf{17.02} & \textbf{1.53}   & \textbf{2.71}  & \textbf{4.90} \\ \bottomrule
\end{tabular}
\label{tab:clean}
\end{table}

\begin{table}[]
\centering
\caption{Comparison on downstream tasks. DF means denoised frames, and IF means interpolated frames. The experiment is taken on Vimeo-90K \cite{xue2019video} dataset.}
\begin{tabular}{@{}lcccc@{}}
\toprule
Model                & DF PSNR        & DF SSIM         & IF PSNR        & IF SSIM         \\ \midrule
RAFT + UNet          & 26.05          & 0.8480          & 24.94          & 0.8266          \\
\textbf{Ours + Unet} & \textbf{26.31} & \textbf{0.8517} & \textbf{25.18} & \textbf{0.8365} \\ \bottomrule
\end{tabular}
\label{tab:downstream}
\end{table}

\begin{table}[]
    \centering
    \caption{The additional verification experiment on HEVC codec.}
    \begin{tabular}{ccc}
    \toprule
         Model &  AEPE & F1\\ \midrule
         Baseline(RAFT)  &  2.50 & 8.71\\
         Ours-MVFlow  &  \textbf{2.40} & \textbf{7.89} \\
    \bottomrule
    \end{tabular}
    \label{tab:hevc}
\end{table}

\end{document}